\documentclass[10pt, letter, onecolumn]{arxiv}
\usepackage[utf8]{inputenc}
\usepackage{textgreek}
\usepackage{kantlipsum, lipsum}
\usepackage{dm-colors}
\usepackage{amsmath}
\usepackage{pstricks, pst-node}
\usepackage{verbatim}
\usepackage{multirow}
\usepackage{rotating}
\usepackage{scalerel}
\usepackage{booktabs}
\usepackage{enumitem}
\usepackage{xspace}
\usepackage{comment}
\usepackage[breakable]{tcolorbox}
\usepackage{bm}
\usepackage{bbm} 
\usepackage{mathtools}
\usepackage{soul}
\usepackage{epsfig}
\usepackage{graphicx}
\usepackage{amssymb}
\usepackage{colortbl}
\usepackage{csquotes}
\usepackage{setspace}
\usepackage{colortbl}
\usepackage{bbding}
\usepackage{threeparttable}
\usepackage{tabularx,ragged2e}
\usepackage{placeins}
\usepackage{bbding}
\usepackage{algorithm}
\usepackage{algorithmic}
\usepackage{fancyvrb}
\definecolor{mainteal}{RGB}{0, 128, 128}
\definecolor{darkpastelgreen}{rgb}{0.13, 0.55, 0.13}
\definecolor{darkpastelred}{rgb}{0.55, 0.13, 0.13}
\usepackage[hang,flushmargin]{footmisc}

\usepackage{nameref}
\usepackage{varioref}
\usepackage{amssymb}%
\usepackage{lineno}
\usepackage{pifont}
\usepackage{xcolor}

\usepackage[pagebackref,breaklinks,colorlinks,
    citecolor=blue,
    filecolor=magenta,
    linkcolor=red,
    urlcolor=cyan]{hyperref}
    
\usepackage[noabbrev,capitalize]
{cleveref}
\usepackage{etoc}

\definecolor{reasoncolor}{HTML}{EE822F}
\definecolor{diagnosecolor}{HTML}{C81D31}
\definecolor{lookupcolor}{HTML}{1E5799}
\definecolor{matchcolor}{HTML}{3B82C4}
\definecolor{searchcolor}{HTML}{5BA3E0}
\graphicspath{{figures/}}
\definecolor{midgreen}{rgb}{0.0, 0.75, 0.0}
\definecolor{softblue}{HTML}{136783}

\newcommand{\ModelName}{TubeMLLM}
\newcommand{\BenchMarkName}{TubeMData}

\definecolor{goodgreen}{HTML}{00B050}
\definecolor{badred}{HTML}{FF0000}

\usepackage{cite}
\usepackage{amssymb}
\usepackage{url}
\usepackage{breakurl}
\usepackage{hyperref}

\usepackage[misc]{ifsym}
\usepackage{bm}
\usepackage{booktabs}
\usepackage{makecell}  
\usepackage{array,color}
\usepackage{array}
\usepackage{multirow}

\usepackage[misc]{ifsym}
\usepackage{multirow}  
\usepackage{booktabs}
\usepackage{threeparttable}
\usepackage{bbding}
\usepackage{ulem}
\usepackage{pifont}
\usepackage{tabularx}

\title{\Large{TubeMLLM: A Foundation Model for Topology Knowledge Exploration in Vessel-like Anatomy}}
\author[1,2]{Yaoyu Liu\textsuperscript{*}} 
\author[1,2]{Minghui Zhang\textsuperscript{*}} 
\author[1,2]{Xin You} 
\author[1]{Hanxiao Zhang} 
\author[1,2,$\dag$]{Yun Gu} 

\affil[1]{\normalsize Institute of Medical Robotics, Shanghai Jiao Tong University, Shanghai, China\\}

\affil[2]{\normalsize Department of Automation, Shanghai Jiao Tong University, Shanghai, China \authorcr \vspace{0.1cm}}

\begin{document}

\begin{abstract}

Modeling medical vessel-like anatomy is challenging due to its intricate topology and sensitivity to dataset shifts. Consequently, task-specific models often suffer from topological inconsistencies, including artificial disconnections and spurious merges. Motivated by the promise of multimodal large language models (MLLMs) for zero-shot generalization, we propose  \textbf{\ModelName}, a unified foundation model that couples structured understanding with controllable generation for medical vessel-like anatomy. By integrating topological priors through explicit natural language prompting and aligning them with visual representations in a shared-attention architecture, TubeMLLM significantly enhances topology-aware perception. Furthermore, we construct TubeMData, a pionner multimodal benchmark comprising comprehensive topology-centric tasks, and introduce an adaptive loss weighting strategy to emphasize topology-critical regions during training.
Extensive experiments on fifteen diverse datasets demonstrate our superiority. Quantitatively, TubeMLLM achieves state-of-the-art out-of-distribution performance, substantially reducing global topological discrepancies on color fundus photography (decreasing the $\beta_{0}$ number error from 37.42 to 8.58 compared to baselines). Notably, TubeMLLM exhibits exceptional zero-shot cross-modality transferring ability on unseen X-ray angiography, achieving a Dice score of 67.50\% while significantly reducing the $\beta_{0}$ error to 1.21. TubeMLLM also maintains robustness against degradations such as blur, noise, and low resolution. Furthermore, in topology-aware understanding tasks, the model achieves 97.38\% accuracy in evaluating mask topological quality, significantly outperforming standard vision-language baselines. 

\end{abstract}

\maketitle

\begingroup
\renewcommand{\thefootnote}{}
\footnotetext{$^*$ Yaoyu Liu and Minghui Zhang contributed equally to this work.}
\endgroup
\begingroup
\renewcommand{\thefootnote}{}
\footnotetext{$\dag$~Corresponding author: Yun Gu. Email addresses: yungu@ieee.org}
\endgroup


\section{INTRODUCTION}

\begin{figure}[!t]
    \centering
    \includegraphics[width=\linewidth]{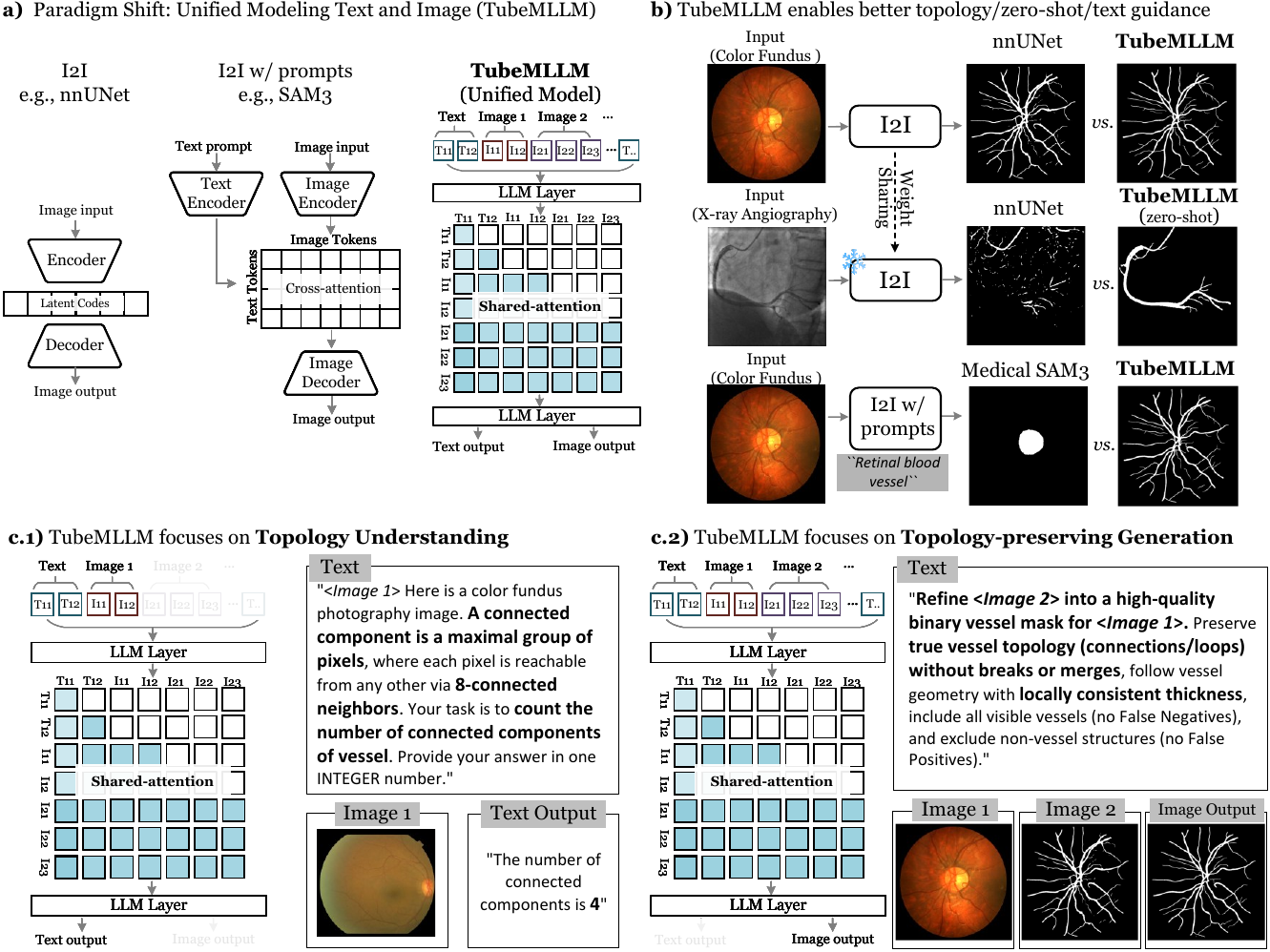}
    \caption{Unified modeling paradigm of the proposed TubeMLLM. (a) Compared with task-specific I2I models and promptable I2I baselines, TubeMLLM unifies text and image tokens in an MLLM with shared-attention and supports both text and image outputs. (b) TubeMLLM enables better topology, stronger zero-shot cross-modality transfer, and more reliable text guidance. (c.1) Example of topology-aware visual understanding task. (c.2) Example of topology-preserving generation task.}
    \label{fig:1}
\end{figure}

Modeling multi-modality medical vessel-like anatomy (e.g., color fundus retinal vasculature and X-ray coronary angiograms) with topology preserved is fundamental to downstream clinical analysis\cite{wang2025research,veneziano2025artificial}, including vascular quantification\cite{sianos2005syntax}, pathology screening\cite{yu2025ensemble}, and intervention planning\cite{veneziano2025artificial}. However, vessel-like structures are inherently difficult because they are thin and elongated, with branching and cyclic connectivity where small local mistakes can cause global topological failures. This challenge is further amplified by dataset shift and modality variations~\cite{litjens2017survey,guan2021domain}. Consequently, existing methods often fail to achieve both high topological fidelity and strong cross-modality transferability.

Specifically, task-specific image-to-image (I2I) segmentation models (e.g., nnUNet~\cite{isensee2021nnu}), illustrated on the left of Fig.~\ref{fig:1}(a), predict masks solely from visual features extracted by an image encoder. To enable topology-preservation in predicted segmentations, substantial efforts have been taken for visual feature enhancement, including architectural design~\cite{isensee2021nnu, liu2022frunet, shi2023afn} and loss functions tailored for tubular structures~\cite{shit2021cldice, kirchhoff2024skeleton,shi2024centerline,zhang2023towards}. Despite these advances, these approaches still exhibit limited generalization under distribution shifts and modality variations, as shown in the
top and middle rows of Fig.~\ref{fig:1}(b).

Recent promptable image-to-image models (I2I w/ prompts)~\cite{carion2025sam3,liu2025medsam3,jiang2026medicalsam3} introduce an independent text encoder to inject language guidance during segmentation, as illustrated in the center of Fig.~\ref{fig:1}(a). However, the input language prompts are typically limited to short phrases such as clinical concepts (e.g., “retinal vessels” in MedicalSAM3~\cite{jiang2026medicalsam3}), which is insufficient to encode complex topology priors such as the definition of connectivity or loops. Furthermore, these models typically output only pixel-level masks. This restricts training primarily to segmentation objectives and prevents the models from leveraging the rich supervision available in language-based understanding tasks~\cite{lai2024lisa,tong2025metamorph,wang2025argenseg}. Consequently, textual guidance for vessel-like structures segmentation is often insufficient. For instance, the bottom row of Fig.~\ref{fig:1}(b) shows a failure case where the optic disc is misclassified as retinal vessels.

To address these limitations, we move beyond the conventional rigid mapping between images, fixed text labels, and pixel-level masks. Inspired by recent MLLM works~\cite{wang2024emu3, deng2025bagel}, we propose TubeMLLM, a unified framework that couples structured understanding with controllable generation for medical vessel-like anatomy. Rather than relying on simple conceptual mappings, TubeMLLM injects explicit topological knowledge through rich, descriptive text prompts, as illustrated in Fig.~\ref{fig:1}(c.1) and Fig.~\ref{fig:1}(c.2) . TubeMLLM simultaneously reasons about the complex topology of vessel-like structures and employs this deep topological understanding to generate faithful outputs. As illustrated on the right of Fig.~\ref{fig:1}(a), TubeMLLM accepts interleaved image and text tokens as input, projects them into a shared feature space, and aligns them through shared-attention within LLM layers. This unified modeling paradigm enables the model to internalize tubular topological priors expressed in rich natural language and associate them with visual features, thereby fundamentally strengthening topology-aware perception.


Building on this unified modeling paradigm, we introduce \textbf{\BenchMarkName}, a pioneer benchmark for topology-aware multimodal medical anatomy learning. As shown in Fig.~\ref{fig:1}(c), TubeMData is a multimodal dataset comprising two synergistic tasks: (i) topological understanding, which analyzes structural properties like connectivity and loops and is formulated as visual question answering; and (ii) topology-preserving generation, which refines or generates vessel-like anatomy with topological consistency and is formulated as image generation. As illustrated in Fig.~\ref{fig:1}(c.1) and Fig.~\ref{fig:1}(c.2), unlike existing promptable image-to-image models, TubeMLLM incorporates both pixel-level mask output and language output. TubeMLLM also replaces short phrases with longer prompts. By explicitly stating imaging modalities as well as topology-related definitions and instructions in the prompts, TubeMLLM is provided with sufficient topology prior guidance.

Further, inspired by recent works that perform additional region-level supervision in image generation\cite{huang2025jova,ma2026text} we incorporate an adaptive loss weighting strategy during training. Specifically, TubeMLLM first establishes spatial correspondence between output pixels and visual tokens, then assigns adaptive weights to each token based on the discrepancy between decoded predictions and ground truth, thereby emphasizing topology-critical and error-prone regions for improved generation performance.

The proposed TubeMLLM is evaluated on fifteen vessel-like datasets spanning two imaging modalities. The experimental results demonstrate the significant improvement in both topological fidelity and segmentation accuracy, effectively reducing topological errors while maintaining robustness against degradations.
\section{RELATED WORKS}
\noindent\textbf{Medical task-specific segmentation models}. Task-specific segmentation models often adopt an image-to-image (I2I) paradigm~\cite{isensee2021nnu,liu2022frunet,shi2023afn,isensee2024nnu}, where visual features extracted from vision encoders are directly used for mask prediction. However, this I2I paradigm is inherently sensitive to dataset distributions, including variations in imaging modalities~\cite{guan2021domain} and anatomical structures~\cite{litjens2017survey}. This limitation becomes particularly evident in vessel-like anatomy segmentation, where thin, elongated tubular structures requires accurate preservation of connectivity and branching patterns. To address this limitation, additional mechanisms are introduced into task-specific models to encourage topology preservation. One line of work focuses on architectural design, enhancing visual representations through specialized network modules~\cite{liu2022frunet,shi2023afn} or dataset-adaptive network configurations~\cite{isensee2021nnu,isensee2024nnu}. Another line of research addresses this issue through loss function engineering, incorporating skeletal structures~\cite{shit2021cldice,zhang2023towards,shi2024centerline,kirchhoff2024skeleton} or persistent homology~\cite{hu2019topology,hu2021topology} into training objectives as priors for topology-aware segmentation. However, these mechanisms still rely on implicit constraints or regularization terms in training, which indirectly enforce topological correctness and therefore provide limited guarantees of topology preservation.

\noindent\textbf{Medical foundation segmentation models}. Foundation segmentation models aim to achieve strong generalization across dataset shifts and often incorporate visual hints (bounding boxes, points) and text prompts as object-level guidance~\cite{kirillov2023sam,zou2023seem,carion2025sam3,lai2024lisa}. Built upon text encoders~\cite{radford2021clip} or large language models (LLM)~\cite{bai2023qwen,touvron2023llama} that are pretrained with vision-language alignment, descriptive linguistic features can be integrated with dense visual features, enabling robust performance across diverse segmentation benchmarks. Bulit on these advances, medical foundation segmentation models~\cite{zhao2025foundation, ma2024segment,ma2025medsam2,zhao2025large,liu2025medsam3,rokuss2025voxtell,jiang2026medicalsam3,xin2025text3dsam} haved been developed. Visual hints or texts are encoded with separate prompt encoders, then fused with visual features using multiplication~\cite{zhao2025large,rokuss2025voxtell} or attention computation~\cite{ma2025medsam2,liu2025medsam3,jiang2026medicalsam3}, to improve the generality of segmentations for various anatomical structures across multiple imaging modalities. 
However, the mask prediction process remains heavily dominated by visual representations, with explicit language guidance either entirely absent~\cite{ma2024segment,ma2025medsam2} or severely constrained to short, rigid phrases~\cite{liu2025medsam3,jiang2026medicalsam3,zhao2025large}. This superficial linguistic guidance precludes the use of rich, descriptive text prompts, significantly limiting its effectiveness in complex anatomical tasks. This limitation is particularly evident in vessel-like anatomy modeling, where maintaining intricate topology (e.g., connectivity and branching) demands strong, explicit topological priors~\cite{hu2019topology,clough2020topological,stucki2023topologically}. 
Furthermore, these models typically rely solely on visual decoders for mask prediction, which restricts training primarily to segmentation objectives. Consequently, they cannot leverage the rich supervision available in language-based understanding tasks~\cite{lai2024lisa,tong2025metamorph,wang2025argenseg}, further increasing their reliance on visual representations.
\section{METHOD}

\begin{figure}[!t]
    \centering
    \includegraphics[width=\linewidth]{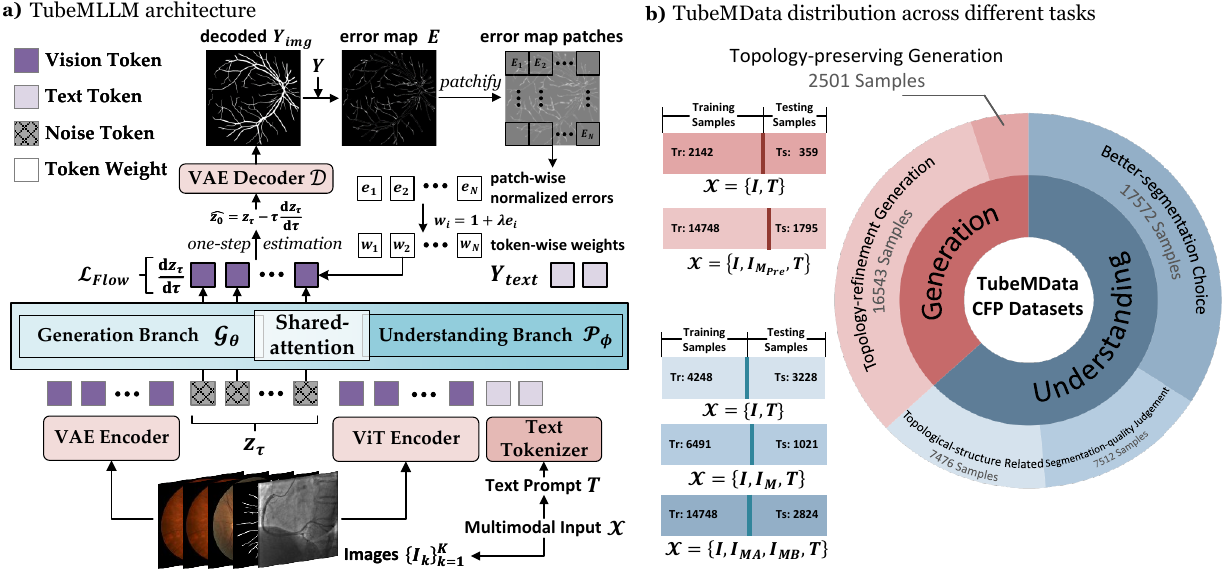}
    \caption{Detailed TubeMLLM architecture and TubeMData. (a) TubeMLLM adopts a Mixture-of-Transformers design with coupled generation transformer branch and understanding transformer branch. Adaptive loss weights are derived from error maps. (b) TubeMData CFP training and testing sample distribution across different topology-centric tasks.}
    \label{fig:2}
\end{figure}

\begin{figure}[!t]
    \centering
    \includegraphics[width=1.0\linewidth]{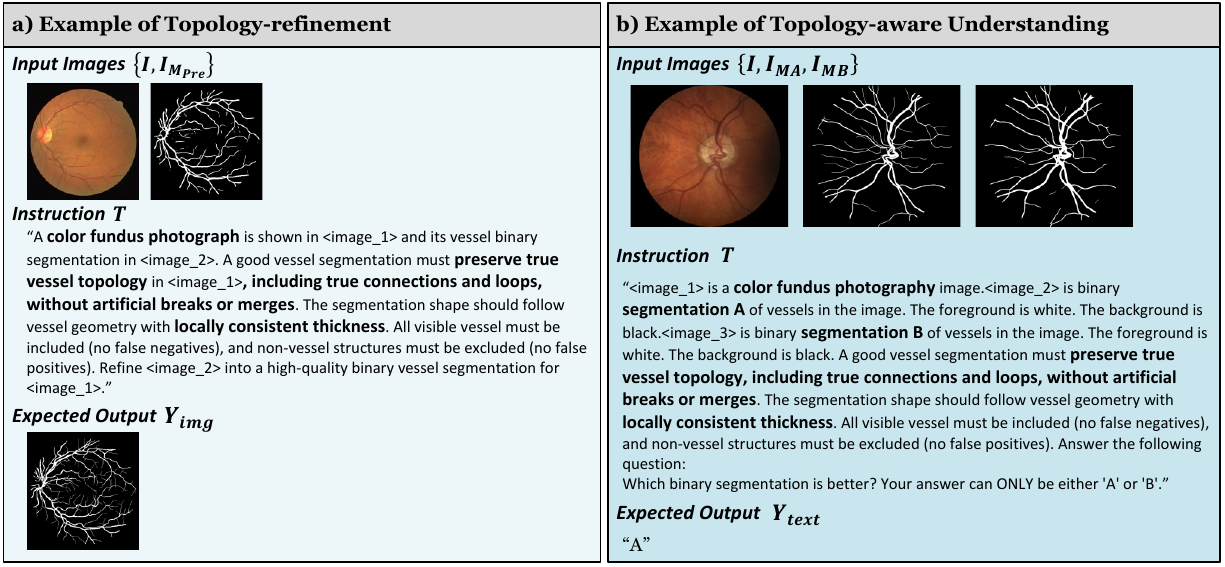}
    \caption{Topology-centric tasks. Bold texts highlight the image modalities and topology priors encoded in language prompts. (a) Topology-refinement generation task. (b) Topology-aware understanding task that select the mask with better topology.}
    \label{fig:2-2}
\end{figure}

\subsection{Problem Formulation and Overall Framework}
The overall architecture of TubeMLLM is illustrated in Fig.~\ref{fig:2}(a). TubeMLLM takes a multimodal input $\mathcal{X}=\big(\{I_k\}_{k=1}^{K},\, T\big)$, where $\{I_k\}_{k=1}^{K}$ denotes the input image set and $T$ is the text prompt, and produces both an image output and a text output, $\mathcal{Y}=\big(Y_{\text{img}},\, Y_{\text{text}}\big)$:
\begin{equation}
    \mathcal{Y} = \mathcal{M}(\mathcal{X})
\end{equation}
where $\mathcal{M}$ denotes TubeMLLM model. Specifically, $Y_{\text{img}}$ is first synthesized in VAE latent space then decoded as:
\begin{equation}
    Y_{\text{img}}=\mathcal{D}(\hat{\mathbf{z}}_{0}) \label{eqn:vae_decode}
\end{equation}
where $\mathcal{D}$ is a frozen VAE decoder and $\hat{\mathbf{z}}_{0}$ is the predicted clean latent. The text output $Y_{\text{text}}$ is generated autoregressively conditioned on $\mathcal{X}$. To support both generation and understanding within a unified framework, TubeMLLM adopts a Mixture-of-Transformers design~\cite{liang2024mot,deng2025bagel} with two coupled branches, $\mathcal{M}=\{\mathcal{G}_{\theta},\, \mathcal{P}_{\phi}\}$. The two branches share joint attention at each layer to enable cross-branch information exchange.
\\ \noindent \textbf{Generation branch.} $\mathcal{G}_{\theta}$ operates on tokenized VAE latents~\cite{labs2025flux1kontextflowmatching} and generates images via rectified flow~\cite{lipman2022flow,labs2025flux1kontextflowmatching}. Image synthesis is formulated as velocity prediction in latent space:
\begin{equation}
    \frac{\mathrm{d}\mathbf{z}_{\tau}}{\mathrm{d}\tau}=\mathcal{G}_{\theta}\!\left(\mathbf{z}_{\tau},\, \tau,\, \mathcal{X}\right),
\end{equation}
where $\mathbf{z}_{\tau}$ denotes the VAE latent at time $\tau\in[0,1]$.
\\ \noindent \textbf{Understanding branch.} $\mathcal{P}_{\phi}$ processes visual tokens extracted by a ViT~\cite{tschannen2025siglip} together with text tokens, and models the conditional distribution of the text output as:
\begin{equation}
    p(Y_{\text{text}}| \mathcal{X}) = \mathcal{P}_{\phi}\left(\mathcal{X}\right)
\end{equation}

\subsection{Adaptive Loss Weighting}
The generative branch $\mathcal{G}_{\theta}$ is trained using the flow-matching objective~\cite{lipman2022flow}, where the predicted velocity $\mathcal{G}_{\theta}\left(\mathbf{z}_{\tau}, \tau, \mathcal{X}\right)$ is aligned with the target velocity $\mathbf{v}_{\text{target}}$ via a mean-squared error (MSE) loss in latent space. To emphasize topology-critical and error-prone regions, we further derive adaptive loss weights from pixel space, as shown in Fig.~\ref{fig:2}(a). Specifically, the predicted clean latent $\hat{\mathbf{z}_0}$ is calculated as:
\begin{equation}
    \hat{\mathbf{z}_0}= \mathbf{z}_{\tau} + (1-\tau)\mathcal{G}_{\theta}\left(\mathbf{z}_{\tau}, \tau, \mathcal{X}\right),
\end{equation}
where $\mathbf{z}_{\tau}$ denotes the noisy latent at time $\tau$. The predicted clean latent $\hat{\mathbf{z}_0}$ is then decoded into $Y_{\text{img}}$ by the frozen VAE decoder $\mathcal{D}$ following Eq.\eqref{eqn:vae_decode}. Given the ground truth image $Y$, the pixel-wise error map is derived as:
\begin{equation}
    E = Y - Y_{\text{img}}    
\end{equation}
As illustrated in Fig.~\ref{fig:2}(a), $E$ is patchified based on the spatial correspondence between visual tokens and image patches in $Y_{\text{img}}$. For the $i$-th visual token ($i=1,2,\dots,N$), we compute within its corresponding patch $E_{i}$ for the normalized mean error intensity $e_{i}$:
\begin{equation}
    e_i = \frac{1}{\left|E_i\right|}\sum_{j\in E_i}\frac{E_i(j)}{E_{\text{max}}}
\end{equation}
where $E_{\text{max}}$ is the maximum possible error intensity within patch $E_i$. The token-level weight $w$ can be derived from the normalized error intensities as:
\begin{align}
    w_{i} &= 1 + \lambda e_{i} \\
    w &= \left\{w_{i}\right\}_{i=1}^{N}
\end{align}
$\lambda$ is set to 10 in all experiments. With token-level weight, the flow-matching loss for generative branch training is formulated as
\begin{equation}
    \mathcal{L}_{\text{Flow}}= \mathbb{E}_{\tau, \mathbf{z}_\tau, \boldsymbol{\epsilon}}
    \left[ \left\| w\odot\big(\mathcal{G}_{\theta}\left(\mathbf{z}_{\tau}, \tau,
    \mathcal{X}\right) - \mathbf{v}_{\text{target}}\big) \right\|^{2}\right].
\end{equation}
This adaptive loss weight formulation emphasizes visual tokens associated with pixel errors and further enhances generation quality with flow matching.

\subsection{Topology-centric Task Design and TubeMData}
\label{subsec:topo-task}
To equip TubeMLLM with topology-preserving generation and topology-aware understanding capabilities, we design a set of topology-centric tasks that explicitly focus on tubular structures. As illustrated in Fig.~\ref{fig:2-2}, these tasks target key topological properties of vessel-like anatomy, including connected components, loops, and the overall topological fidelity of a predicted mask. Based on these topology-centric tasks, we construct TubeMData dataset, part of which is illustrated in Fig.~\ref{fig:2}(b).

\noindent\textbf{Topology-preserving Generation}. For topology-preserving generation, we introduce topology-refinement task, as shown in Fig.~\ref{fig:2-2}(a). $\mathcal{X}=\left\{I, I_{\text{M}_{\text{Pre}}}, T\right\}$ includes the original image $I$ and a preliminary prediction $I_{\text{M}_{\text{Pre}}}$ with imperfect topology. Conditioned on the topology constraints specified in the instruction $T$, the model $\mathcal{M}$ generates a refined mask $Y_{\text{img}}$ that better preserves topological consistency.

\noindent\textbf{Topology-aware Understanding}. For topology-aware understanding, we design VQA tasks and encode topology priors through query prompts $T$, as illustrated in Fig.~\ref{fig:1}(c). These tasks operate at two input granularities. In the first setting, the input is either $\mathcal{X}=\left\{I, I_{\text{MA}}, I_{\text{MB}}, T\right\}$ with two candidate masks $I_{\text{MA}}, I_{\text{MB}}$ (Fig.~\ref{fig:2-2}(b)), or $\mathcal{X}=\left\{I, I_{\text{M}}, T\right\}$ with a single mask $I_{\text{M}}$. The model produces $Y_{\text{text}}$ to (i) select the mask with better topology, or (ii) judge whether the mask satisfies topology requirements. In the second setting, $\mathcal{X}=\left\{I, T\right\}$ contains only the image $I$, and the model $\mathcal{M}$ predicts in $Y_{\text{text}}$ the existence or the number of certain topological structures (e.g., connected components, loops). 

These topology-centric tasks take advantage of the shared-attention design in TubeMLLM via interleaved image-text inputs and provide supervision from both image generation ($Y_{\text{img}}$) and textual prediction ($Y_{\text{text}}$), thereby strengthening topology-aware representation learning.

\noindent\textbf{TubeMData Construction}. Fig.~\ref{fig:2}(b) illustrates the distribution of TubeMData color fundus photography (CFP) samples with different generation and understanding tasks, as well as the training and test split within each task. Specifically, ten public CFP datasets as well as five public X-ray angiography (XRA) datasets are included in TubeMData\footnote{\textbf{CFP}: FIVES~\cite{FIVES}, RETA~\cite{RETA}, RVD~\cite{RVD}, STARE~\cite{STARE} for training; AFIO~\cite{AFIO}, ARIA~\cite{ARIA}, CHASEDB1~\cite{CHASEDB1}, DRIVE~\cite{DRIVE}, IOSTAR~\cite{IOSTAR}, LES-AV~\cite{LESAV} for testing. \textbf{XRA}: DCA1~\cite{DCA1}, Dr-SAM~\cite{DrSAM}, FS-CAD~\cite{FSCAD}, XCAV~\cite{XCAV} for training; XCAD~\cite{XCAD} for testing.}.  To assess cross-domain generalization, the test split is strictly out-of-distribution from the training split. The aggregated corpus contains approximately 3.5K images with heterogeneous spatial resolutions and domain distributions. Topological-imperfect predictions used in the topology-centric tasks are generated using nnUNet~\cite{isensee2021nnu} models trained with different loss functions~\cite{shit2021cldice, kirchhoff2024skeleton,shi2024centerline, zhang2023towards}. For topology-aware understanding tasks, ground-truth answers are automatically derived via rule-based verification of topological structures in both manual labels and imperfect masks. Overall, TubeMData contains approximately 52K samples.

\section{EXPERIMENTS}
\makeatletter \def\hlinew#1{%
\noalign{\ifnum0=`}\fi\hrule \@height #1 \futurelet \reserved@a\@xhline} \makeatother
\begin{table}[!t]
    \renewcommand{\arraystretch}{1.2}
    \caption{Results on CFP OOD datasets. T-T denotes topology-centric tasks.}
    \centering
    \label{tab:1}\scalebox{1.0}{
    \begin{tabular}{lcccc}
        \hlinew{1pt} \textbf{Methods}               & \textbf{Dice}$\uparrow$ & \textbf{clDice}$\uparrow$ & \textbf{$\beta_{0}$ Num}$\downarrow$ & \textbf{$\beta_{0}$ Mat}$\downarrow$ \\
        \hlinew{1pt} nnUNet~\cite{isensee2021nnu}   & 74.44                   & 78.48                     & 37.42                                & 36.56                                \\
        nnUNet w/ clDice~\cite{shit2021cldice}      & 75.19                   & 79.83                     & 25.98                                & 27.62                                \\
        nnUNet w/ SR~\cite{kirchhoff2024skeleton}   & 74.76                   & 79.68                     & 34.49                                & 30.52                                \\
        nnUNet w/ cbDice~\cite{shi2024centerline}   & 74.94                   & 79.34                     & 25.46                                & 30.11                                \\
        nnUNet w/ CAL~\cite{zhang2023towards}       & 72.63                   & 77.85                     & 56.91                                & 29.79                                \\
        \hlinew{0.5pt} FR-UNet~\cite{liu2022frunet} & 71.59                   & 76.76                     & 40.13                                & 27.94                                \\
        AFN~\cite{shi2023afn}                       & 60.35                   & 59.62                     & 24.29                                & 34.65                                \\
        \hlinew{0.5pt} SAM3~\cite{carion2025sam3}   & 0.33                    & 0.27                      & 11.08                                & 29.11                                \\
        MedicalSAM3~\cite{jiang2026medicalsam3}     & 12.14                   & 10.46                     & 10.33                                & 30.74                                \\
        Bagel~\cite{deng2025bagel}                  & 30.68                   & 38.01                     & 307.14                               & 178.09                               \\
        \hlinew{0.8 pt} TubeMLLM w/o T-T            & 75.73                   & 80.43                     & \textbf{8.47}                        & 26.57                                \\
        \textbf{TubeMLLM}                           & \textbf{76.09}          & \textbf{80.59}            & 8.58                                 & \textbf{18.99}                       \\
        \hlinew{1pt}
    \end{tabular}}
\end{table}

\makeatletter \def\hlinew#1{%
\noalign{\ifnum0=`}\fi\hrule \@height #1 \futurelet \reserved@a\@xhline} \makeatother
\begin{table}[!t]
    \renewcommand{\arraystretch}{1.2}
    \caption{Comprehensive quantitative results including refinement, cross-dataset,
    and low quality input settings. ZS denotes zero-shot setting, FS denotes from sratch setting. GB, GN, LR denotes Gaussian Blur, Gaussian Noise, and Low Resolution.}
    \centering
    \label{tab:2} \scalebox{1.0}{
    \begin{tabular}{lccccc}
        \hlinew{1.2pt} \textbf{Task} & \textbf{Method} & \textbf{Dice}$\uparrow$ & \textbf{clDice}$\uparrow$ & \textbf{$\beta_{0}$ Num}$\downarrow$ & \textbf{$\beta_{0}$ Mat}$\downarrow$ \\
        \hlinew{1.2pt}                                                                                                                            
        \multicolumn{6}{l}{\textit{Task: Single Step Refinement(SSR)}}                                                                         \\
        \hlinew{0.5pt} \multirow{2}{*}{SSR} & nnUNet~\cite{isensee2021nnu} & 74.44 & 78.48 & 37.42 & 36.56 \\
        \cline{2-6}                         & w/ TubeMLLM                 & \textbf{76.05} & \textbf{80.43} & \textbf{8.49}  & \textbf{26.51} \\
        \hlinew{0.8pt}                                                                                                                            
        \multicolumn{6}{l}{\textit{Task: Cross Dataset}}                      \\
        \hlinew{0.5pt} \multirow{2}{*}{ZS}  & nnUNet~\cite{isensee2021nnu} & 9.07  & 11.74 & 238.26 & 5.29  \\
        \cline{2-6}                         & TubeMLLM                     & \textbf{67.50} & \textbf{69.17} & \textbf{1.21}   & \textbf{7.69}  \\
        \hlinew{0.3pt} \multirow{2}{*}{FS}  & nnUNet~\cite{isensee2021nnu} & 77.08 & 75.43 & 12.06  & 0.81  \\
        \cline{2-6}                         & TubeMLLM                     & \textbf{77.51} & \textbf{78.67} & \textbf{1.31}   & \textbf{0.69}  \\
        \hlinew{0.8pt}                                                                                                                            
        \multicolumn{6}{l}{\textit{Task: Low Quality Image Input Setting}} \\
        \hlinew{0.5pt}
        \multirow{2}{*}{GB}                 & nnUNet~\cite{isensee2021nnu} & 78.04 & 81.98 & 26.57 & 21.44 \\
        \cline{2-6}                         & TubeMLLM                     & \textbf{81.13} & \textbf{84.91} & \textbf{1.93}  & \textbf{12.34} \\
        \hlinew{0.3pt} \multirow{2}{*}{GN}  & nnUNet~\cite{isensee2021nnu} & 74.77 & 71.82 & 18.36 & 24.10 \\
        \cline{2-6}                         & TubeMLLM                     & \textbf{77.30} & \textbf{73.51} & \textbf{1.68}  & \textbf{13.90} \\
        \hlinew{0.3pt} \multirow{2}{*}{LR}  & nnUNet~\cite{isensee2021nnu} & 77.92 & 81.17 & 23.61 & 21.43 \\
        \cline{2-6}                         & TubeMLLM                     & \textbf{80.25} & \textbf{83.53} & \textbf{1.68}  & \textbf{12.32} \\
        \hlinew{1.2pt}
    \end{tabular}}
\end{table}

\begin{figure}[!t]
    \centering
    \includegraphics[width=1.0\linewidth]{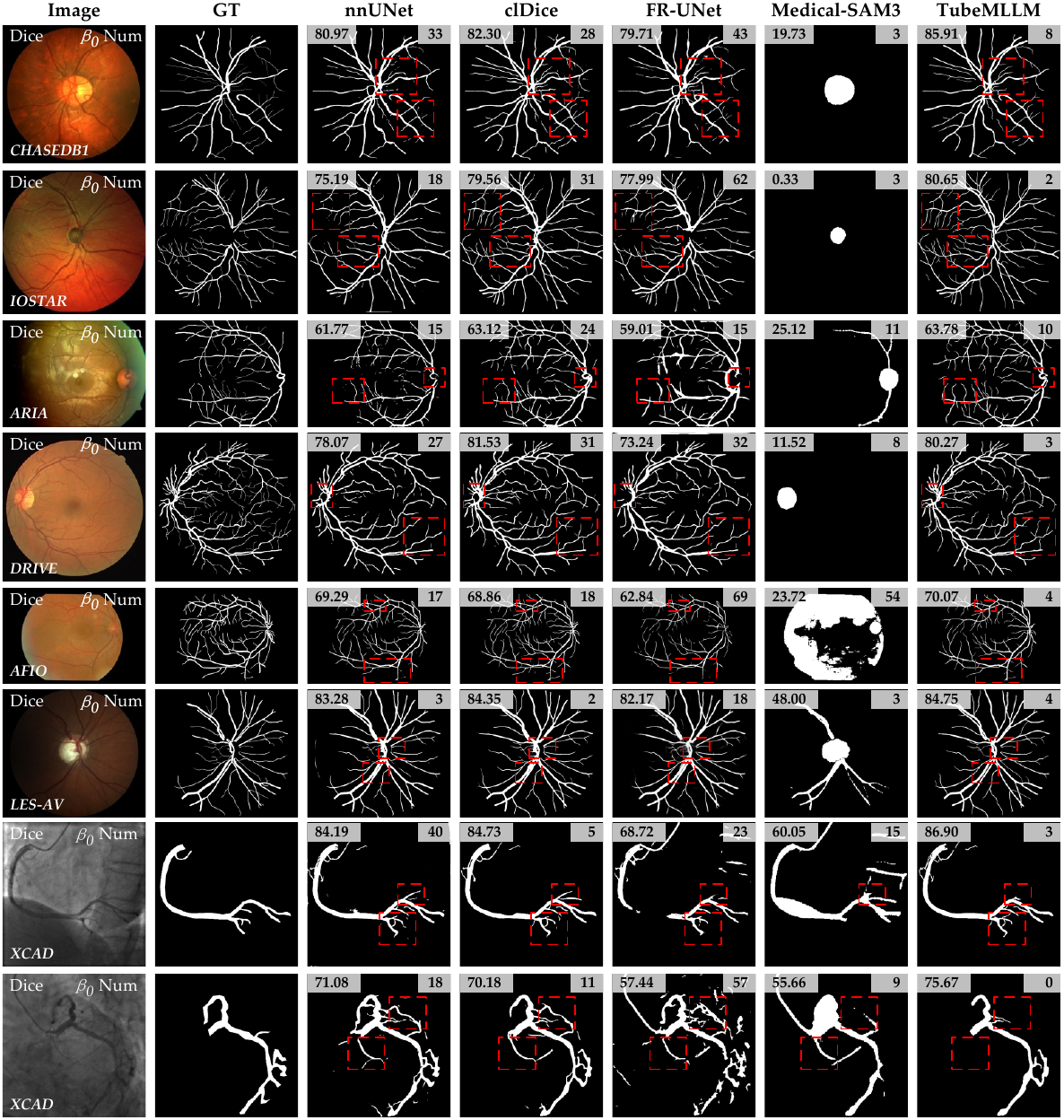}
    \caption{Qualitative results on CFP and XRA OOD test datasets. $\beta_0$ \textrm{Num} denotes the $\beta_0$ number error. Regions inside red boxes are highlighted to demonstrate the topological accuracy of TubeMLLM.}
    \label{fig:3}
\end{figure}

\begin{figure}[!t]
    \centering
    \includegraphics[width=0.85\linewidth]{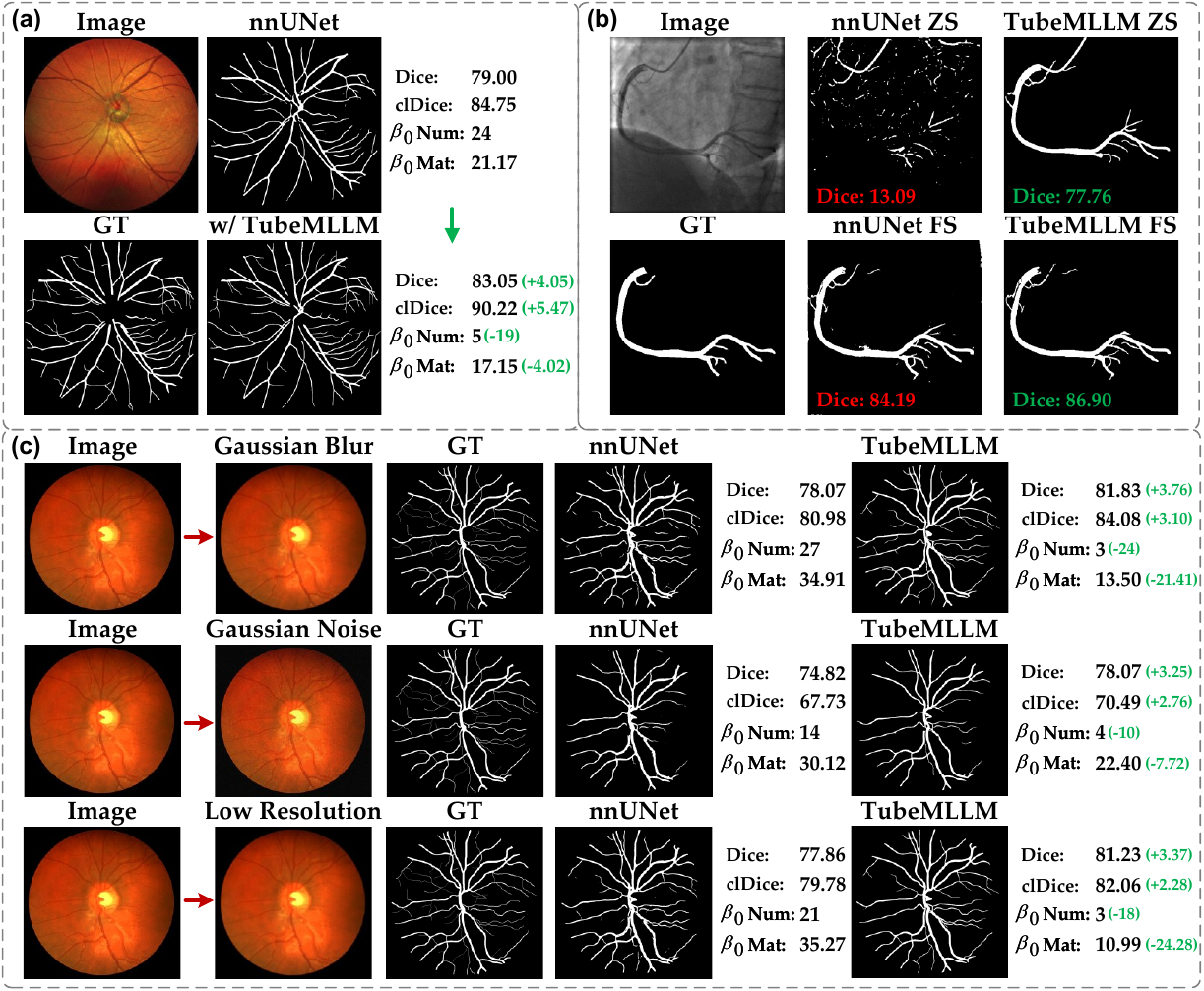}
    \caption{Qualitative performance on (a) topology-refinement, (b) zero-shot transfer to XRA dataset and (c) degraded input of CFP datasets. Text colored in green demonstrates superior results or improved performance.}
    \label{fig:4}
\end{figure}

\subsection{Implementation Details}
During training, TubeMLLM was finetuned on TubeMData from the pretrained model\cite{deng2025bagel}. For evaluating the generation branch, we reported $\beta_0$ number error ($\beta_0$ \textbf{Num}) and $\beta_0$ matching error~\cite{stucki2023topologically} ($\beta_0$ \textbf{Mat}) to quantify global and local topological discrepancies. For understanding evaluation, we used mean absolute error for topological-structure counting, and classification accuracy for the remaining tasks.

\subsection{Topology-preserving Generation}
Tab.~\ref{tab:1} evaluates segmentation performance across three distinct paradigms: topology-enhanced nnUNet variants~\cite{isensee2021nnu, shit2021cldice, kirchhoff2024skeleton, shi2024centerline, zhang2023towards}, specialized vessel architectures~\cite{liu2022frunet, shi2023afn}, and promptable foundation models~\cite{carion2025sam3, jiang2026medicalsam3, deng2025bagel}. It can be observed that TubeMLLM outperforms all baselines in Dice, clDice, and topological metrics. Specifically, TubeMLLM reduces the $\beta_{0}$ number error to 8.58 and the $\beta_{0}$ matching error to 18.99, a substantial improvement over the $\sim$30 error range typical of nnUNet-based variants. This gap underscores the efficacy of our topology-explicit prompting in enforcing structural consistency. While foundation models like SAM3 and MedicalSAM3 achieve competitive $\beta_{0}$ counts, their markedly lower Dice and clDice indicate that they fail to capture the overall structure of vessel-like anatomy. Fig.~\ref{fig:3} provides qualitative visual comparisons where TubeMLLM generates fewer topological errors as highlighted in red boxes.

In addition, we evaluated TubeMLLM on topology-refinement, cross-modality transfer, and zero-shot robustness under low-quality inputs, as summarized in Tab.~\ref{tab:2}. The results show that TubeMLLM substantially enhances the topological fidelity of imperfect segmentation masks, reducing the $\beta_{0}$ number error from 37.42 to 8.49. Notably, TubeMLLM also exhibits strong zero-shot generalization to the XRA modality, achieving a Dice score of 67.50 while decreasing the $\beta_{0}$ number error from 238.26 to 1.21. Fig.~\ref{fig:4}(b) and Fig.~\ref{fig:tubemllm-zs-XRA} in the appendix further illustrate this improvement. Under low-quality input settings, Tab.~\ref{tab:2} indicates that TubeMLLM remains robust in terms of topological fidelity, outperforming nnUNet by approximately 3\% in Dice and reducing the $\beta_{0}$ number error by more than 20 across all three degradation scenarios.

\begin{figure}[!t]
    \centering
    \includegraphics[width=1.0\linewidth]{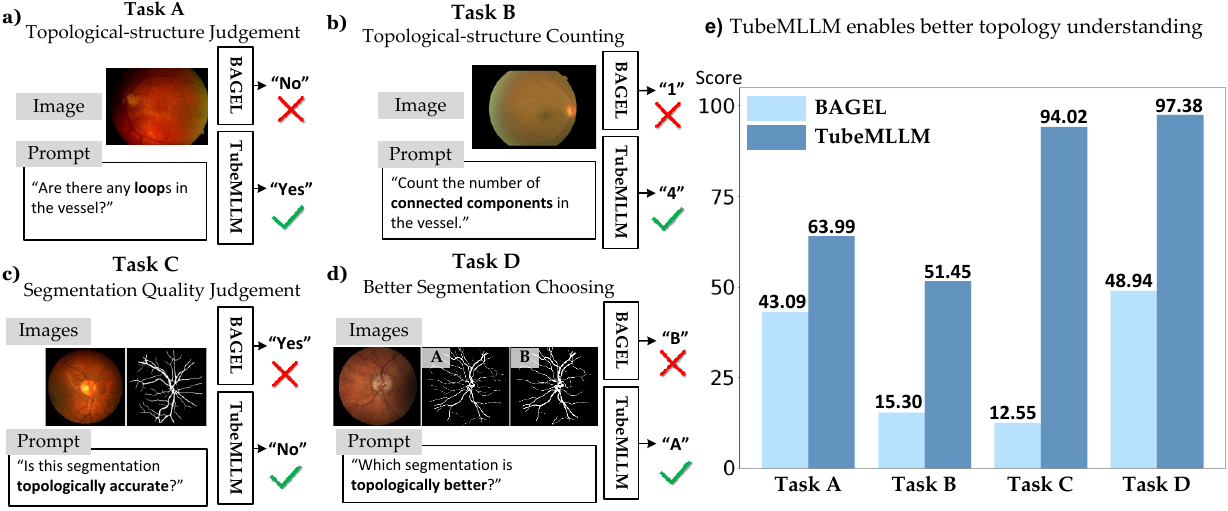}
    \caption{Qualitative and quantitative performance on four understanding tasks. (a-d) Qualitative comparisons between BAGEL and our TubeMLLM on (a) Topological-structure Judgement, (b) Topological-structure Counting, (c) Segmentation Quality Judgement, and (d) Better Segmentation Choosing. (e) Quantitative results demonstrating that TubeMLLM significantly outperforms the baseline across all four tasks.}
    \label{fig:5}
\end{figure}

\begin{figure}[!t]
    \centering
    \includegraphics[width=1.0\linewidth]{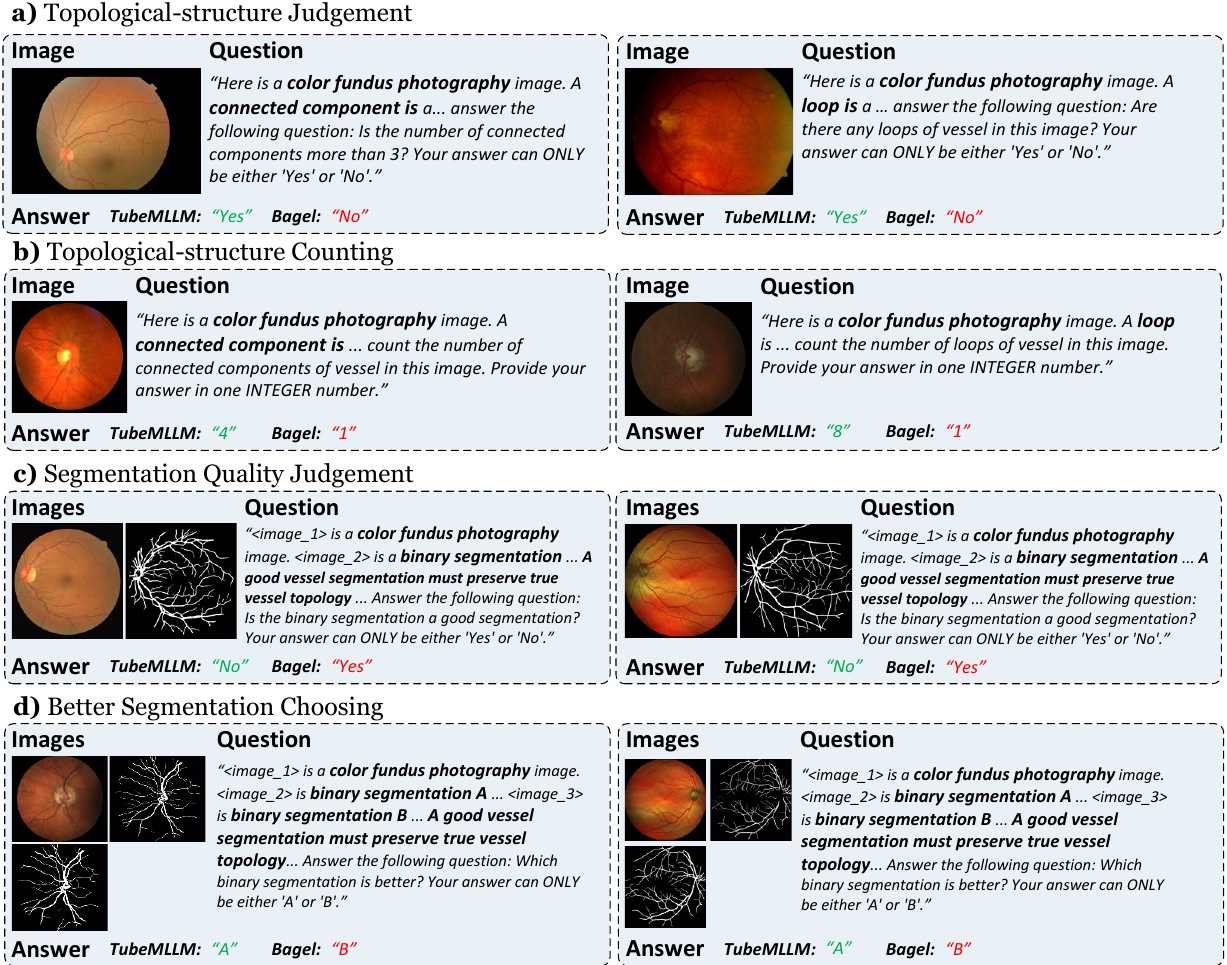}
    \caption{Illustration of TubeMLLM performance on topology-aware understanding tasks. Texts colored in green represent correct answers. Texts colored in red represent incorrect answers. Bold texts highlight the image modalities and topology priors encoded in language prompts.}
    \label{fig:tubemllm-und}
\end{figure}

\subsection{Topology-aware Understanding} 
Fig.~\ref{fig:5}(a-d) and Fig.~\ref{fig:tubemllm-und} present qualitative results of topology understanding tasks. It can be seen that TubeMLLM has a better understanding of vessel topology after finetuning on TubeMData. Specifically, TubeMLLM makes accurate judgement on the existence of connected components and loops and correctly counts the number of these structures in OOD CFP images, as shown in Fig.~\ref{fig:5}(a,b) and Fig.~\ref{fig:tubemllm-und}(a,b). In Fig.~\ref{fig:5}(c,d) and Fig.~\ref{fig:tubemllm-und}(c,d), TubeMLLM also demonstrates the capability of evaluating segmentation mask quality based on topology. Fig.~\ref{fig:5}(e) further shows that TubeMLLM quantitatively improves the ability to count and classify topological structures. Moreover, TubeMLLM achieves 97.38\% accuracy in distinguishing good from poor topological quality, a substantial gain over the baseline (48.94\%), demonstrating its effectiveness in assessing overall topological quality.

\section{CONCLUSION}
We propose TubeMLLM, a unified foundation model that integrates vessel-like anatomical priors through language and aligns them with visual representations to enhance topology-preserving perception. Trained on the comprehensive TubeMData with a topology-centric task design and adaptive loss weighting, TubeMLLM substantially improves both topological fidelity and segmentation accuracy, while demonstrating robust generalization to out-of-distribution datasets. Overall, TubeMLLM opens new possibilities for modeling vessel-like anatomy with topological fidelity in a unified multimodal framework.

\clearpage
\balance
\bibliographystyle{unsrt} 
\bibliography{references} 

\clearpage
\appendix
\section{APPENDIX}
\setcounter{table}{0}
\renewcommand{\thetable}{A\arabic{table}}
\setcounter{figure}{0}
\renewcommand{\thefigure}{A\arabic{figure}}
\setcounter{equation}{0}
\renewcommand{\theequation}{A\arabic{equation}}
\subsection{Details of TubeMData Benchmark}
\begin{figure}[!t]
    \centering
    \includegraphics[width=0.7\linewidth]{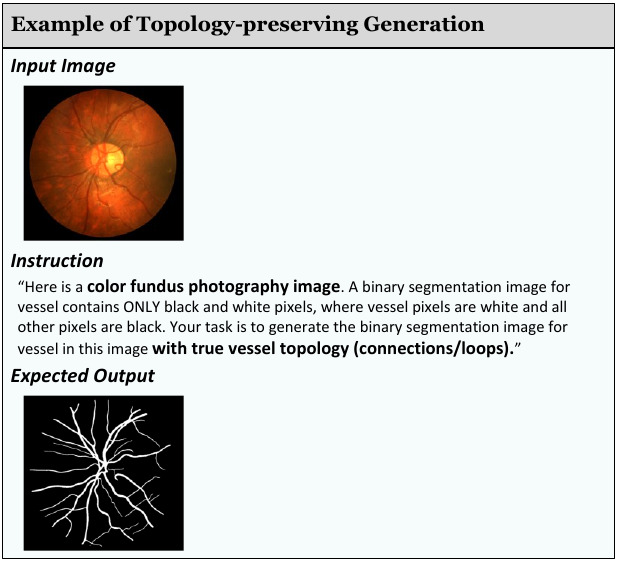}
    \caption{Example of topology-preserving generation. Bold texts highlight the image modalities and topology priors encoded in language prompts.}
    \label{fig:topo-preserving-refinement}
\end{figure}

\begin{figure}[!t]
    \centering
    \includegraphics[width=0.7\linewidth]{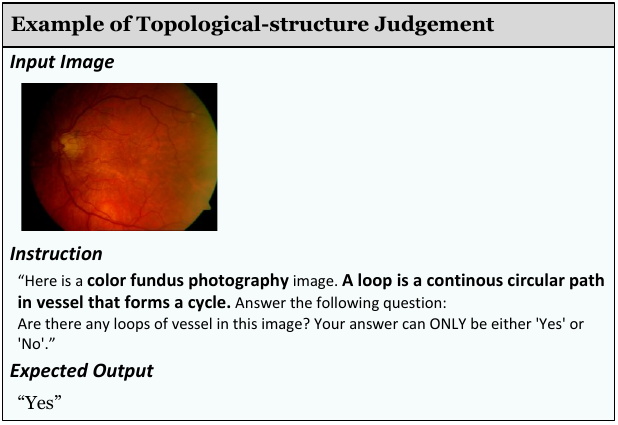}
    \caption{Example of topological-structure judgement. Bold texts highlight the image modalities and topology priors encoded in language prompts.}
    \label{fig:topo-structure-judgement}
\end{figure}

\begin{figure}[!t]
    \centering
    \includegraphics[width=0.7\linewidth]{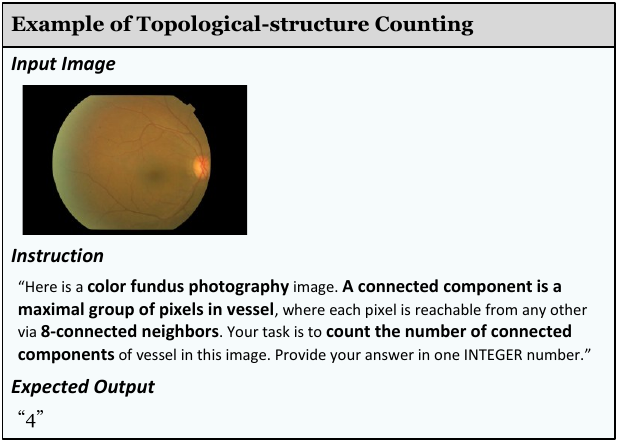}
    \caption{Example of topological-structure counting. Bold texts highlight the image modalities and topology priors encoded in language prompts.}
    \label{fig:topo-structure-count}
\end{figure}

\begin{figure}[!t]
    \centering
    \includegraphics[width=0.7\linewidth]{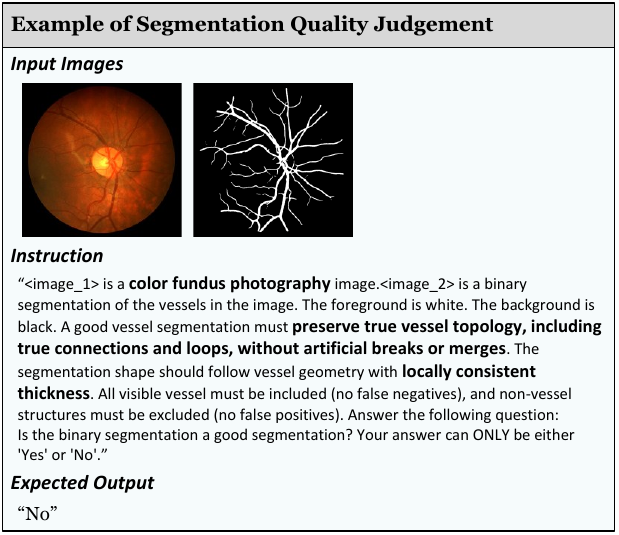}
    \caption{Example of segmentation topology-quality judgement. Bold texts highlight the image modalities and topology priors encoded in language prompts.}
    \label{fig:seg-quality-judgement}
\end{figure}

Fig.~\ref{fig:topo-preserving-refinement} to Fig.~\ref{fig:seg-quality-judgement} illustrate an example for topology-preserving generation task, topological structure judgement, topological structure counting and segmentation quality judgement, respectively. Definitions of topological structures and criteria of good segmentation are inserted into the language prompts as topology priors. These topology-centric tasks take advantage of the shared-attention design in TubeMLLM via interleaved image-text inputs and provide supervision from both image generation ($Y_{\text{img}}$) and textual prediction ($Y_{\text{text}}$), thereby strengthening topology-aware representation learning.
\subsection{TubeMLLM Generalization on XRA}
\begin{figure}[!t]
    \centering
    \includegraphics[width=\linewidth]{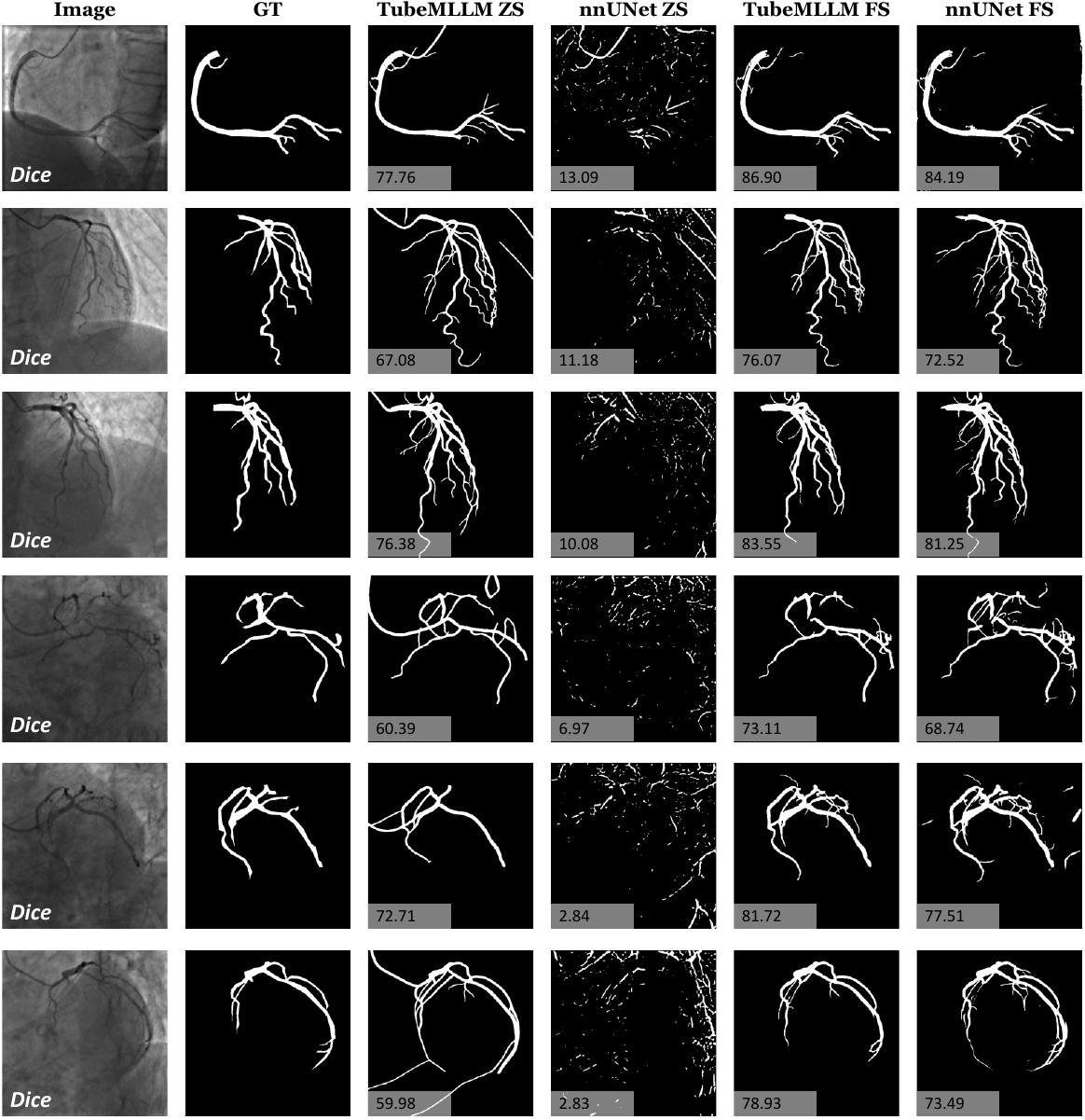}
    \caption{Qualitative results of TubeMLLM zero-shot generalization on XRA. ZS denotes zero-shot setting, FS denotes from sratch setting.}
    \label{fig:tubemllm-zs-XRA}
\end{figure}

Fig.~\ref{fig:tubemllm-zs-XRA} shows the results of TubeMLLM zero-shot generalization and finetuned adaptation to XRA modality. It is evident that TubeMLLM is capable of delineating the vessel structure in unseen XRA images even without finetuning, achieving $\sim$50 dice improvement compared with nnUNet models. Even after finetuning, TubeMLLM still generates XRA vessel masks with better accuracy and topology. These results further demonstrate the zero-shot generalization capability of TubeMLLM.

\newpage

\end{document}